\documentclass[sigconf]{acmart}
\AtBeginDocument{%
  }

\setcopyright{acmlicensed}
\copyrightyear{2025}
\acmYear{2025}
\acmConference[ICAIL '25]{Make sure to enter the correct
  conference title from your rights confirmation email}{June 03--05,
  2018}{Woodstock, NY}

\usepackage{svg}
\usepackage{listings}
\usepackage{xcolor}
\usepackage{caption}

\begin{document}

\pagenumbering{arabic}

\title{Are Manual Annotations Necessary for Statutory Interpretations Retrieval?}

\author{Aleksander Smywiński-Pohl}
\orcid{0000-0001-6684-0748}
\email{apohllo@agh.edu.pl}
\affiliation{%
  \institution{AGH University}
  \city{Krakow}
  \country{Poland}
}

\author{Tomer Libal}
\orcid{0000-0003-3261-0180}
\email{tomer.libal@uni.lu}
\affiliation{%
  \institution{University of Luxembourg}
  \city{Belval}
  \country{Luxembourg}}

\author{Adam Kaczmarczyk}
\orcid{0009-0003-8171-4987}
\email{adam.kaczmarczyk@uni.lu}
\affiliation{%
  \institution{University of Luxembourg}
  \city{Belval}
  \country{Luxembourg}}

\author{Magdalena Król}
\orcid{0000-0003-0392-0921}
\email{magdakrol@agh.edu.pl}
\affiliation{%
  \institution{AGH University}
  \city{Krakow}
  \country{Poland}
}
\renewcommand{\shortauthors}{Smywiński-Pohl et al.}

\begin{abstract}
One of the elements of legal research is looking for cases where judges have extended
the meaning of a legal concept by providing interpretations of what a concept
means or does not mean. This allow legal professionals to use such interpretations
as precedents as well as laymen to better understand the legal concept.

The state-of-the-art approach for retrieving the most relevant interpretations for these concepts
currently depends on the ranking of sentences and the training of language models over annotated examples.
That manual annotation process can be quite expensive and need to be repeated for each such concept,
which prompted recent research in trying to automate this process. In this paper, we highlight the results
of various experiments conducted to determine the volume, scope and even the need for manual annotation.

First of all, we check what is the optimal number of annotations per
a legal concept. Second, we check if we can draw the sentences for annotation
randomly or there is a gain in the performance of the model, when
only the best candidates are annotated. As the last question
we check what is the outcome of automating the annotation process with
the help of an LLM.
\end{abstract}

\ccsdesc[500]{Information systems~Retrieval models and ranking}
\ccsdesc[300]{Applied computing~Annotation}

\keywords{Statutory Interpretation, Ranking Algorithms, Large Language Models, Annotation Process}

\maketitle

\section{Introduction}

Statutory Interpretation (SI) is an important task in the scope of Artificial Intelligence (AI) and Law.
It is concerned with the interpretation of legal concepts made by judges in the court rulings. 
The classical example is the sentence ``No vehicles in the park'' with respect to the \textit{vehicles}
concept. Does it cover bicycles and ambulances?

A possible help from an AI system for such a task could offer a quick access to the judgments or
fragments thereof containing statements that provide such interpretations. A system could identify
a sentence appearing in one of the judgments, e.g. ``Although a bicycle is commonly classified as a vehicle,
the purpose of the sign is to keep the park as a quiet place, so bicycles are exempt from it.'' as
highly relevant for interpreting what a \textit{vehicle} is in that context.

The problem of finding such sentences could be treated as a general legal information retrieval and theoretically
addressed with the state-of-the-art retrieval models based on deep neural networks, such as BGE \cite{bge2024chen}.
However, the current task is special in the sense that we want to exclude a large group of texts found in the judgments,
i.e. those excerpts that only cite the regulation without providing any extended interpretation of the legal concept.
Due to the fact that such models are based on similarity of the query and the searched text, such excerpts would 
appear at the top of search results, when such a model is applied. So this task requires a more specific model
to be solved correctly.

Creating a specific model requires building a properly designed dataset, which contains information about the utility of the explanatory sentences retrieved by a general retriever. The construction of such a dataset is not a trivial task, as it requires annotators with a high level of expertise. The annotators are supposed to discover the true meaning of retrieved sentences in the context of legal explanation and decide whether the examples are useful or not in the task. This process is complicated and understanding of the sentences can be subjective and be prone to some other factors such as annotator fatigue \cite{savelka2020discovering}. The framework needed to mitigate the issues mentioned above appears to be very costly, as it must cover many factors that can have a potential impact on the interpretation of legal semantics.

In this research, we want to check how the annotation process, necessary to train the model, can be optimized in order
to improve its cost effectiveness. Namely, we have conducted various experiments to try and answer different questions regarding this process, with the goal of creating some initial optimization guidelines.

We highlight the results of three experiments which answer some of these questions and outline a basic guide.

Our first experiment tries to answer the question "given a statutory concept, {\bf how many examples do we need to annotate before the results converge (RQ1)?}".

Second, we want to check "{\bf how the choice of specific sentences to annotate might affect the quality of the model (RQ2)?}".

Lastly, we check "{\bf to what extent we can use a language model to annotate the examples (RQ3)?}", following research by \cite{savelka2023can}.

In the experiments, we tested these questions over different versions of an open source language model. The answers to these three questions allows us to write some initial guidelines to help with future SI retrieval tasks. Our benchmark is  the state-of-the-art results regarding SI retrieval obtained in \cite{smywinski2024enhancing}. The benchmark studies used different reranking methods to assess the usefulness of the exemplary sentences.

In the next section, we overview the current state-of-the-art in data annotation, retrieval-augmented generation, and transfer learning in the context of law. This builds on the foundational research on argument retrieval presented in \cite{ashley2013information}, establishing the first approach for building argument-retrieval systems - the systems enabling the SI, as well as recent advancements that demonstrate the capabilities of large language models in specialized legal tasks documented by \cite{bommarito2023gpt} and later \cite{Katz2024gpt4}. We then present our method, which closely follows the one introduced by Savelka's PhD Thesis \cite{savelka2020discovering}, in which he used the model to first find the explanatory examples and later to rank them with 4 categories of usefulness. This is followed by our description of the three experiments and then by the results of these experiments and detailed guidelines for SI annotations in different scenarios. We conclude the results and provide a summarized guide to help SI retrieval tasks.


\section{Related Work}
\label{sec:sota}

As the subject of this article is optimizing the annotation process for statutory interpretation, 
this section presents the literature on various aspects concerning this process. Among others, they are:
\begin{itemize}
    \item facilitating annotation \textit{per se},
    \item limiting the burden of labeling, or increasing the effectiveness of those efforts by using transfer learning, as well as
    \item retrieval of the best candidates to be included in a dataset.
\end{itemize}

The most relevant research, i.e. the Ph.D. of Jaromir Savelka \cite{savelka2021discovering} is discussed in section \ref{sec:method}.

\subsection{Data Annotation in the Context of Law}

The problem of annotation in the context of law has been addressed by Gray et al. \cite{gray2023can}. 
They used an LLM (\verb|gpt-3.5-turbo-16k|) to preannotate sentences of legal opinions in Drug-Interdiction Auto-Stop (DIAS) cases. The annotation of the sentences indicated which DIAS factor (if any) is present in a sentence. The LLM was provided with the annotation instruction and dozens of examples of proper annotation, both included in the prompt. However, in the described solution, the LLM was not an independent annotator but only an assistant that proposed the labels to human annotators. It constituted efficient support in the annotation process by making the work faster while not negatively influencing the outcome of the process, nor making the experts completely rely on it. The LLM hints increased the time efficiency of the annotation process.

Savelka et al. \cite{savelka2023can} used LLMs in the annotation task requiring highly specialized domain expertise. 
The task here was to label sentences from court opinions to explain legal concepts (the same dataset we use in this research).
\verb|GPT-4| model was provided with annotation guidelines in the prompt. It occurred that the model performed similarly to well-trained student annotators, maintaining good quality even for batch predictions. The level of annotator agreement (Krippendorff's $\alpha$) of the LLM was in the middle of the pack of all annotators (LLM and human).

It is worth mentioning that the literature seems to not undertake the topic of number of training examples, which is a part of the research presented in this article.

\subsection{Retrieval-Augmented Generation in the Context of Statutory Data}
In their recent work Luo at al. \cite{unknown} present ATRI, a retrieval-augmented generation framework for interpreting vague legal concepts using past judicial precedents, alongside the new Legal Concept Entailment benchmark for automated evaluation, demonstrating that the system’s outputs effectively assist large language models and are comparable to expert-written interpretations.

Concentrating on multi-layered system, de Oliviera Lima \cite{DBLP:journals/corr/abs-2411-07739} proposed embedded texts—generating dense vector representations for individual articles, their subdivisions (paragraphs, clauses), and structural groupings (books, titles, chapters) in order to capture hierarchical complexities and enhance information retrieval. His aim was to demonstrate broad applicability to different legal systems and other domains with similarly structured text.

Savelka and Ashley \cite{vsavelka2022legal} provide foundational work on legal information retrieval focused on statutory interpretation, outlining methods for discovering sentences relevant to statutory terms and illustrating the limitations of traditional keyword-based techniques in this context.

\subsection{Transfer Learning in the Context of Law}
The data and the importance of reusing existing models was thoroughly examined by Savelka et al. \cite{10.1145/3462757.3466149} where functional segmentation of the judgments in cross-contextual and cross-jurisdictional tasks were revised and described. Researchers used language-agnostic sentence embeddings in sequence labeling models using Gated Recurrent Units (GRUs) to investigate transfer between different contexts and jurisdictions in the task of functional segmentation of adjudicatory decisions. The examined models appeared to be able to generalize beyond the context (e.g. jurisdiction) they had been trained on, as well as to be more robust and to have better overall performance during the evaluation of previously unseen contexts.

Tyss et al.\cite{tyss2024beyond} transferred pre-trained models for legal case summarization to jurisdictions without available reference summaries for training, underlying the role of pre-training in the case summarization problem. Nevertheless, the choice of the dataset for pretraining should be based on lexical and jurisdictional similarity rather than its size or abstractiveness, which shows that the transfer cannot be performed discretionarily. 

Savelka et al.\cite{savelka2021cross} used transfer learning to predict rhetorical roles in different domains and jurisdictions, demonstrating the ability of language models to generalize and abstract beyond the specific domain vocabulary. The article also shows that training the models on pools of data taken from different datasets can leverage their performance and robustness. A similar dataset augmenting approach is also presented in the paper discussed next.

Joel Niklaus et al.\cite{niklaus2022empirical} improved the performance of the BERT family models in the task of predicting legal judgment by augmenting the training dataset with case law from different jurisdictions. The model trained on the augmented datasets showed better performance than those trained on data from singular jurisdiction owing to, as the authors of the described paper believe, informational gain from other diverse cases. Unlike in the previous approaches, the transfer did not occur \textit{per se}, but by transferring the information through the dataset augmentation and performing a single extended training process.

Furthermore, to emphasize elder approaches and the longer presence of cross-jurisdictional transfer in the AI and law domain, Savelka and Ashley\cite{vsavelka2015transfer} used statistical machine learning to classify specific functional categories in statutory texts from multiple jurisdictions in US states. The transfer of the statistical model helped to solve the problem of sparse data and its imbalance in different jurisdictions. The transfer improved the classification results.

Zheng et al. \cite{zheng2021pretraining} emphasized the conditions under which pre-training improves performance on legal tasks, identifying key factors such as data similarity and structure. Chalkidis et al. \cite{chalkidis2020legalbert} demonstrated that domain-specific pretraining of BERT on legal corpora significantly enhances performance on a wide range of downstream legal tasks.

The literature mentioned above indicates that transfer learning is an efficient way of improving the models' performance in various tasks, especially when the data is sparse or lacking, as it reduces the effort of data gathering and annotation.

\section{Approach}
\label{sec:method}

This research concerns the task of discovering sentences for argumentation about the meaning of statutory terms. This task, 
introduced by \v{S}avelka and Ashley, is defined in \cite{vsavelka2022legal} as a specific type of legal argument retrieval, by 
itself defined by Ashley and Walker \cite{ashley2013information} as the merging of legal information retrieval and legal argument mining.

\subsection{The Dataset}

Savelka in his Ph.D. thesis \cite{savelka2020discovering} constructed
a dataset of 42 concepts with more than 27 thousand sentences scored with 
respect to their value regarding statutory interpretation. These sentences were retrieved from the Caselaw access 
project\footnote{https://case.law} and where selected according to the sentences containing specific occurrences of 
a list of chosen legal concepts. For their experiment, they have chosen to fine-tune the language model RoBERTa-base, which is 
a pre-trained transformer-based language model developed by Facebook (currently Meta) AI. 
These findings were  reviewed in \cite{smywinski2024enhancing} where the authors have shown that a 
better performance for this task can be obtained with a DeBERTa v.3 model \cite{he2021debertav3} with a voting scheme.

The task considered by \v{S}avelka and Ashley is to return, given legal concept and a provision, a list of sentences 
which best explain the legal meaning of the concept. Such sentences can be  definitional sentences 
\cite{vsavelka2022legal} that state explicitly in a different way what the statutory phrase means
or state what it does not mean, by providing an example, instance, or counterexample of the phrase, 
and sentences that show how a court determines whether something is such an example, instance, or counterexample.
In order to be able to train and assess the models, the dataset was annotated by law students where each 
sentence was annotated by two students. The annotators assigned each sentence a category denoting whether it 
has a \textit{high}, \textit{certain}, \textit{potential} or \textit{no value} to understanding the legal concepts.

In order to evaluate the quality of their fine-tuned models, the researchers used Normalized Discounted Cumulative Gain score, 
very popular in information retrieval  \cite{jarvelin2002cumulated}.  
For the task of argument mining they first defined $S_j=(s_1,\ldots,s_n)$, where $s_i$ for $0<i\leq n$ is a sentence for concept $j$ in the $i$-th
place in the list of retrieved sentences. They then used, for the purpose of assigning a value for each $S_j$, and for a given $k$, a 
normalized discounted cumulative gain as follows:
\begin{equation}
    NDCG(S_j,k) = \frac{1}{Z_{jk}}\Sigma_{i=1}^k\frac{rel(s_i)}{log_2(i+1)}
\end{equation}
where $rel(s_i)$ is the value of each sentence for the understanding of a concept ($3$ for \textit{high} value to $0$ for \textit{no value}) 
and $Z_{jk}$ normalizes the result by dividing it with the value of the ideal sorting of the sentences. 
The reader is invited to consult with \cite{jarvelin2002cumulated} for a detailed explanation of this measure.

\subsection{The Task Data Format}

The data in the core of the interest of this article are the sentences extracted from case law paired with the legal concept that they explain. The legal concepts are extracted from statutory law in a process of legal analysis. The examples used for demonstration are from the case law of the European Patent Office Board of Appeal\footnote{https://www.epo.org/en/legal/case-law}. A data point contains two fields relevant from the point of view of the research: the text of the sentence and the legal concept attached (Fig. \ref{fig:datapoint-initial}).

\begin{figure}[h]

\centering
\begin{minipage}{.8\linewidth}
\begin{lstlisting}[language=Python,
    basicstyle=\ttfamily\small,
    frame=single,
    breaklines=true,
    breakatwhitespace=true,
    columns=flexible
]
{
    "text": "Thus, a chemical compound can involve an inventive step irrespective of whether it itself has an unexpected technical effect, or whether its effect is linked to the improvement in a complete processing, as is the case for the improvement in Z-isomer yield directly attributable to the intermediate compound (1) of claim 1, as set out above.",
    "concept": "involvesInventiveStep"
}
\end{lstlisting}
\end{minipage}
\captionof{figure}{Example of a data point in JSON format used for annotation.}
\label{fig:datapoint-initial}
\end{figure}

After annotation, the example is given a label which expresses its explanatory value for the legal concept (Fig. \ref{fig:datapoint-annotation}): 

\begin{figure}[h]

\centering
\begin{minipage}{.8\linewidth}
\begin{lstlisting}[language=Python,
    basicstyle=\ttfamily\small,
    frame=single,
    breaklines=true,
    breakatwhitespace=true,
    columns=flexible
]
{
    ...,
    "value": "high"
}
\end{lstlisting}
\end{minipage}
\captionof{figure}{Example of a data point in JSON format after annotation.}
\label{fig:datapoint-annotation}
\end{figure}

The model predicting the explanatory value instead of returning a discrete value returns a continuous value, which is a measure that allows to find the most relevant sentences (Fig. \ref{fig:datapoint-prediction}.

\begin{figure}[h]

\centering
\begin{minipage}{.8\linewidth}
\begin{lstlisting}[language=Python,
    basicstyle=\ttfamily\small,
    frame=single,
    breaklines=true,
    breakatwhitespace=true,
    columns=flexible
]
{
    ...,
    "value": "2.8547831177711487"
}
\end{lstlisting}
\end{minipage}
\captionof{figure}{Example of a data point in JSON format after model prediction.}
\label{fig:datapoint-prediction}
\end{figure}

\subsection{The Training Setup}

For the purpose of fine-tuning their best model, they have used examples pairing the sentences with the 
provision of the legal concepts, where the provisions were defined as the smallest text in the regulation expressing a statutory
provision regarding the legal concept.
Lastly, they have divided their data into 6 folds, 4 of which were used for training in a 4-fold cross-validation 
setup (in each training one of the folds is used as 
the evaluation fold and 3 remaining folds are used to train the model) and 2 for final testing. In order to ensure proper 
distribution of the data among the folds, they have classified each legal concept into one of four categories and 
ensured the same number of elements of each category are in each fold.

This research was extended by Smywiński-Pohl and Libal in \cite{smywinski2024enhancing}. The authors have tested 
a number of additional models and settings for sorting the sentences, using the same dataset as the input.
The authors have found out that DeBERTa v. 3 \cite{he2021debertav3} in the \textit{large} variant gives the best 
results for this task. 


\begin{table}[ht]
\centering
\begin{tabular}{c|c|c}
    voting     & NDCG@10   & NDCG@100   \\ \hline
no  & 0.790  & 0.786 \\ \hline
yes & \textbf{0.817}  & \textbf{0.801} \\
\end{tabular}
\caption{Results obtained by the models presented in \cite{smywinski2024enhancing} on the test subset.}
\label{tab:smywinski}
\end{table}

Their best result of running the model on the test for $k=10$ and $k=100$ is summarized in  Table \ref{tab:smywinski}. 
The table concerns two setups: one without a voting scheme and the other including the voting scheme. Since
the training follows a cross-validation procedure with 4 models, it is possible to use all these models to
decide on the final score of the sentence. We present the best results with and without the voting scheme,
while in the second setup the score is the average score obtained by all models on the test set.

In the following experiment we will use the approach presented in \cite{smywinski2024enhancing} and \cite{savelka2023can}
to answer the research questions. For the first two experiments we will train a cross-encoder model following
closely the training paradigm presented in \cite{smywinski2024enhancing} and we will present the averaged
scores obtained on the test splits. For the last experiment we will follow the LLM approach presented
in \cite{savelka2023can} but we will extend the results for the full test set. This will enable us to present
the NDCG scores and compare the different approaches directly.

\section{Research Questions}

\subsection{RQ1: How many examples should be manually annotated?}


The number of annotated examples required for effective model training was not addressed in \cite{savelka2021discovering}. As noted in that work, annotation is time-consuming and error-prone. Our first research question thus explores whether only a subset of available sentences needs to be annotated to achieve comparable results.

This topic also connects with broader literature on cost-effective annotation strategies. Ein-Dor et al. \cite{ein-dor2020active} investigate how few training examples are required when employing active learning for BERT-like models. Their work supports the hypothesis that significant gains can be achieved with fewer annotations if selection is optimized.

Our approach to answer this question is to randomly pick up to $k$ sentences for each concept, with 
values for $k$ ranging from 100 to 1000 with 100 steps. In case the number of available sentences
for a concept is below $k$, we take the whole set for this concept. The model trained over
the chosen sentences is then compared to a model which was trained on all sentences. 



In this research question we assume that for each concept we
take up to $k$ sentences, to train the models and to compute their performance
on the full testing subset.
We compare the result with the performance of the models trained on the whole dataset.
This means that for some concept all examples will be taken, if the total number of 
sentences found is lower than $k$. The assumption is that $k$ is the same for all concepts,
even though there is a different distribution of sentences for different concepts and a different
distribution of labels for a given concept. We left the question how to adapt the 
number of examples depending on the sentence and value distribution for future research. 
For $k$ tested in the experiment we take values from 100 to 1000 with 100 step. We take 
1000 as the maximum, since in the preliminary experiment we have observed that there is almost no
difference between training on up to 1000 examples for each concept and the full training set.


\subsection{RQ2: Which sentences to choose for annotation?}

Our second research question focuses on the sentences chosen for annotation and whether
using a preliminary sorting of the sentences can provide a better result than a random choice.

As a reminder, the sentences are classified according to four different classes, with an unknown distribution among the sentences.

Our approach to answer this question has two items. First, we made a decision to consider as higher quality for the purpose of training, sentences which are classified as more relevant for giving an interpretation for the concept. Second, we decided to use active learning in order to achieve that.

This method is informed by prior work on active learning and sentence selection, including Gray et al. \cite{gray2023can} and Westermann et al. \cite{10.1145/3462757.3466149}, which show that LLMs and embeddings can assist in prioritizing high-value sentences.

Active learning is a process where training examples are used selectively and incrementally to train a model. Our approach to select the best candidates is to use a previously trained model to rank and sort the sentences. For each of the four splits under consideration, a model is created by training on the other annotated three splits.

The iterative and incremental element comes from the request of the algorithm, from the user, to annotate specific examples at each phase, which are ranked according to the model from the previous iteration. In this way, the accuracy of choosing sentences which are classified as most relevant increases at each step.

In our experiment, we even take a step further and consider a model trained on all examples of the three remaining splits, without the iterative and incremental phases. The rational behind that is to answer the question whether this approach, given optimal settings, brings a value to the annotation process. We have therefore taken an optimal setting by considering all examples of the other splits as training data points.

Although this approach cannot be 
reproduced in practice, due to lack of already annotated examples, it is quite useful, as will be shown later, to answer what can be achieved with such an approach.

The model in the preliminary sorting phase is used as follows. First, the model is applied in
order to rank each of the sentences. We then sort the sentences according to this rank, where
more relevant sentences appear first. We then repeat the experiment from RQ1, but this time,
the sentences are taken from the sorted list and are not chosen randomly.

\subsection{RQ3: Do we need the manual annotation at all?}

In the last question we want to check if the whole annotation effort is necessary at all.
Thus we follow the approach presented by Savelka et al in \cite{savelka2023can} and use an LLM as the annotator.

Similarly to \cite{savelka2023can}, \cite{bommarito2023gpt}, \cite{Katz2024gpt4} we create a prompt based on the same annotation guidelines and we pass each sentence together
with the concept and the provision to the LLM and ask it to provide an annotation label.

There are several differences between our approach and that taken in \cite{savelka2023can}. First, in \cite{savelka2023can}, only 256 sentences were automatically selected. This is because of the cost associated with using GPT4, a closed-code model. Our use of an open source one allowed us to automatically annotate all the 11k examples in the test set. Thus, obtaining a more accurate estimation of the model.

Second, the fact that we have automatically annotated all examples in the test set allowed us to go beyond checking only the accuracy and F1 score of the model, as is done in \cite{savelka2023can}. Since our goal is to provide the 10 or 100 most relevant sentences, the NDCG scores are more relevant than accuracy of F1 scores.

Besides the label of each class, we also register the probabilities associated with first tokens compatible with
the valid labels. This allows us not only to compute the classification scores such as 
accuracy and F1, but also to compute the real-value score of the sentence and sort 
the sentences according to that score. This will allow to apply this method directly
for presenting the top sentences to the end user.

\section{Results}

To answer RQ1 and RQ2 we apply the following training procedure. Like \cite{savelka2020discovering}, we train 4 models in a cross-validation setup,
so each model is trained on 3 splits and validated during training on the remaining split. Following \cite{smywinski2024enhancing} we use
DeBERTa (v. 3) base (184 million parameters) and large (435 million parameters) variants
and use the hyper-parameters given in table \ref{tab:hyperparams}. We save the model after every epoch and compute the NDCG@10 score (\textit{metric}
parameter in the table) on
the validation set to select the best model for a given training, which is the used to compute the NDCG@10 and NDCG@100 scores on the testing set.
We use 4 labels, even though according to the findings in \cite{smywinski2024enhancing} the number of labels could be reduced to just 2.
We train for 5 epoch with a batch size of 8. There is no warm up, we apply a linear decay schedule for the learning rate, which starts from 2e-05
and we use 768 as the maximum number of tokens that is passed to the model, even though the input can be longer. In the text passed to the model
for classification, we put the concept first, then the sentence to be assessed and the provision at the end. So in the case the input
is too long, the provision might be shortened. These trainings are repeated 5 times with random generator seed set to ${0, 1, 2, 3, 4}$.
We repeat these training for each tested value of $k$ (100, 200, 300 \ldots), so we get 200 trainings in total (10 values of $k$, 4 splits,
5 repeats for different random seeds) for each model size and each setup (RQ1 and RQ2).

\begin{table}
    \centering
    \begin{tabular}{l r}
    \hline
    \textbf{Parameter} & \textbf{Value} \\
    \hline
    metric & NDCG@10 \\
    labels & 4 \\
    FP16 & true \\
    LR & 2e-05 \\
    epochs & 5 \\
    batch size & 8 \\
    warm up ratio & 0 \\
    max length & 768 \\
    \hline
    \end{tabular}
    \caption{Hyper-parameters used for training the base and large variants of the DeBERTa v. 3 models.}
    \label{tab:hyperparams}
\end{table}

\begin{figure}
    \centering
    \includegraphics[width=0.95\linewidth]{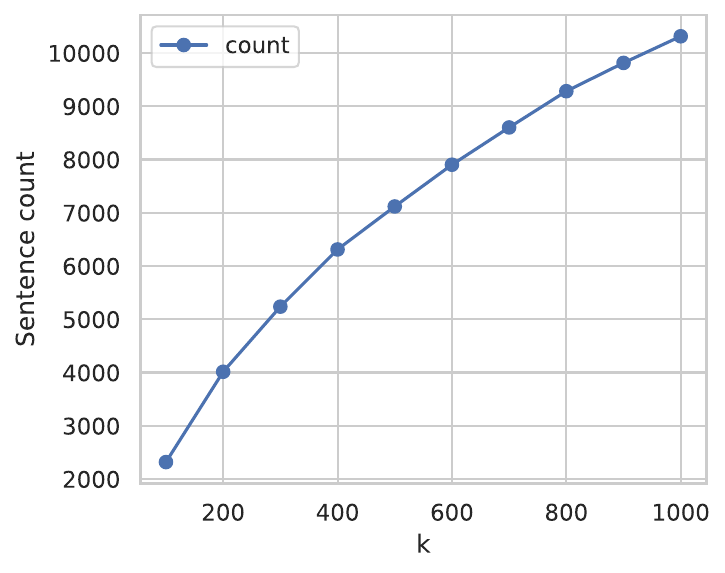}
    \caption{The total number of sentences in the training subset for different values of the maximum number of sentences
    ($k$) taken for each concept.}
    \label{fig:counts}
\end{figure}

Figure \ref{fig:counts} contains the total number of sentences in the training subset (i.e. four splits) for different
values of $k$. The scaling is not linear, since the distribution of sentences in the dataset is not even. Since the deltas
for growing values of $k$ are getting smaller, we see that the number of concepts having at least $k$ sentences is shrinking
quickly.

\subsection{RQ1: How many examples should be manually annotated?}

\begin{figure*}
    \centering
    \includegraphics[width=0.95\linewidth]{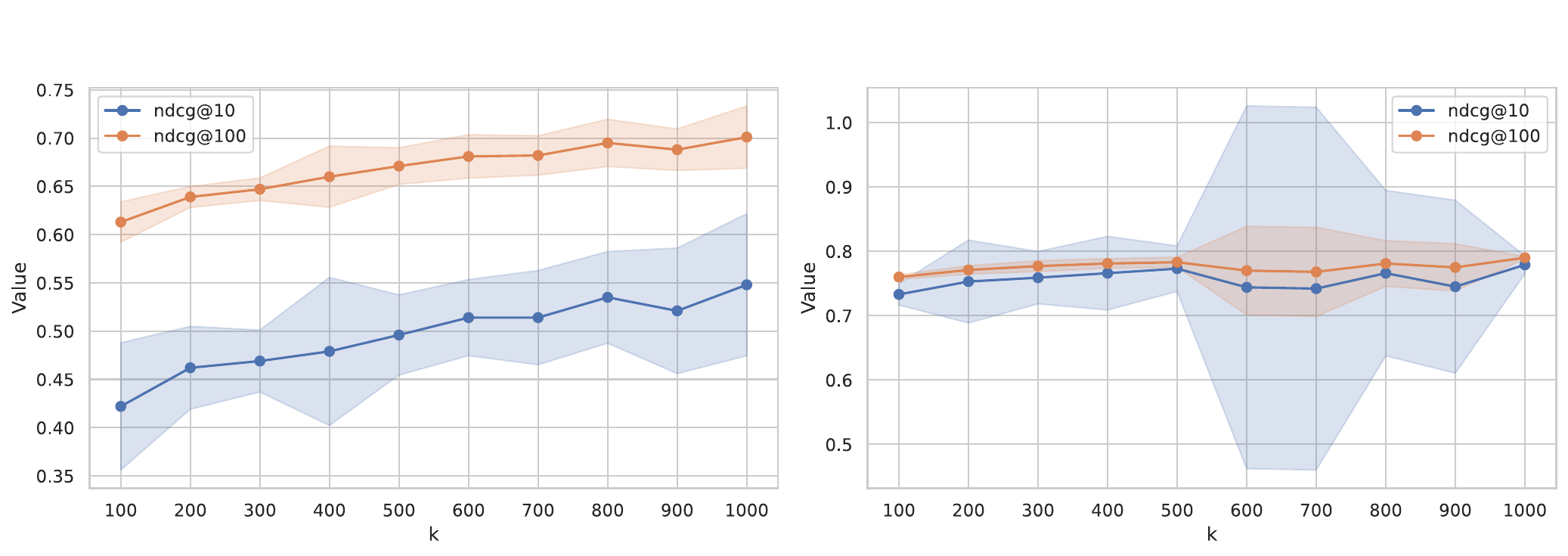}
    \caption{NDCG@10 and NDCG@100 results for the DeBERTa \textit{base} (left) and \textit{large} (right) model trained on top-k \textit{random} sentences.
    The shaded contour indicates the standard deviation of the scores for 5 runs.}
    \label{fig:shuffle}
\end{figure*}

The results for the first experiment are given in Table \ref{tab:shuffle} and Figure \ref{fig:shuffle}.
We report the scores obtained on the test subset averaged among the splits and different values of the random seed.
On the figure we plot the standard deviation of the results for different values of the random seed (we take the
averages of the splits as the input to compute the variability among the different seeds). For the large
variant of the model for some values of $k$ we can observe a huge standard deviation among the results.

We compare the \textit{base} model with the \textit{large}
model and  the results differ significantly for these  setups. For the base
model we observe that with a growing values of $k$ the performance of the model grows.
We get 0.42 NDCG@10 for k=100 and 0.55 for k=1000, which is a +13 pp. improvement.
For NDCG@100 it grows from 0.61 up to 0.70, a +9 pp. improvement. We also observe that
the we could gain even more from a larger number of examples in this setup, since
for the full training dataset we have an NDCG@10 score of 0.68 (+13 pp. improvement
compared to k=1000) and NDCG@100 score of 0.76 (16 pp. improvement) which are definitely
large differences.

\begin{table}
    \centering
    \begin{tabular}{c c c c c}
        \hline
        & \multicolumn{2}{c}{\textbf{Base}} & \multicolumn{2}{c}{\textbf{Large}} \\
        \hline
    \textbf{Top-k} & @10 & @100 & @10 & @100 \\ \hline
    100  & 0.422 & 0.613 & 0.733 & 0.760 \\
    200  & 0.462 & 0.639 & 0.753 & 0.771 \\
    300  & 0.469 & 0.647 & 0.759 & 0.777 \\
    400  & 0.479 & 0.660 & 0.766 & 0.781 \\
    500  & 0.496 & 0.671 & 0.773 & 0.783 \\
    600  & 0.514 & 0.681 & 0.744 & 0.770 \\
    700  & 0.514 & 0.682 & 0.742 & 0.768 \\
    800  & 0.535 & 0.695 & 0.766 & 0.781 \\
    900  & 0.521 & 0.688 & 0.745 & 0.775 \\
    1000 & 0.548 & 0.701 & 0.779 & \textbf{0.790} \\ \hline
    \textbf{full} & \textbf{0.681} & \textbf{0.761} & \textbf{0.790} & 0.786 \\ \hline

    \end{tabular}
    \caption{NDCG scores @10 and @100 of training DeBERTa (v.~3) \textit{base} and \textit{large} models on up to $k$ random sentences for each concept.}
    \label{tab:shuffle}
\end{table}

The outcome for the large model is very different, i.e. the performance improvement
thanks to the growing number of examples is very small. We reach a peak score of 77.3\% NDCG@10 for
500 examples and then 77.9\% NDCG@10 for 1000 examples (compared to 79.0\% for the
full dataset, a -1.2 pp. difference). For NDCG@100 there's also a peak at 500 examples (78.3\%) and
the best score is for 1000 examples (79.0\%) which is better than the result for the full
dataset (78.6\%). So we either observe small differences (like 1 pp. drop in NDCG@10 if we limit the
number of examples to 500) or even improvements with a smaller number of examples (NDCG@100 for
1000 is better than for the full dataset). The second phenomenon might be due to the fact that
without the threshold, one of the concepts in the training dataset dominates it, and the models
overfit to that concept, loosing their generalization power.

To summarize these findings we can conclude that if we are going to train a small model
(since e.g. we are concerned with the deployment costs) we should annotate as many examples
as possible. However for the larger model we can limit the number of examples to 500
(if we accept a 1 pp. drop in performance) or 1000 which could even bring us improvements
compared to the training on the full dataset.

\subsection{RQ2: Which sentences to choose for annotation?}

The second research questions concerns the problem of selecting the sentences for annotation.
In RQ1 we took a random sample of sentences containing the legal concept. Here we take a different
approach -- we take top-$k$ examples, not a random sample. To sort these sentences for each
split in the training subset we take a model trained on the remaining splits including the full
training dataset. This setup would make no much sense in the real setting, i.e. when we are
building a new dataset for statutory interpretation. But this experiment should be viewed
primarily as an optimistic limit for a setup when such a model is available. Here we think
specifically on the transfer learning scenario, i.e. how much annotation effort we could save
(in the best case) if we took a dataset or a model trained for statutory interpretation
in different jurisdiction or on a completely different set of legal concepts. Since we cannot
expect that in such a case we could obtain better results than in this experiment we can
use this information to make our choices regarding the annotation process.

\begin{figure*}
    \centering
    \includegraphics[width=0.95\linewidth]{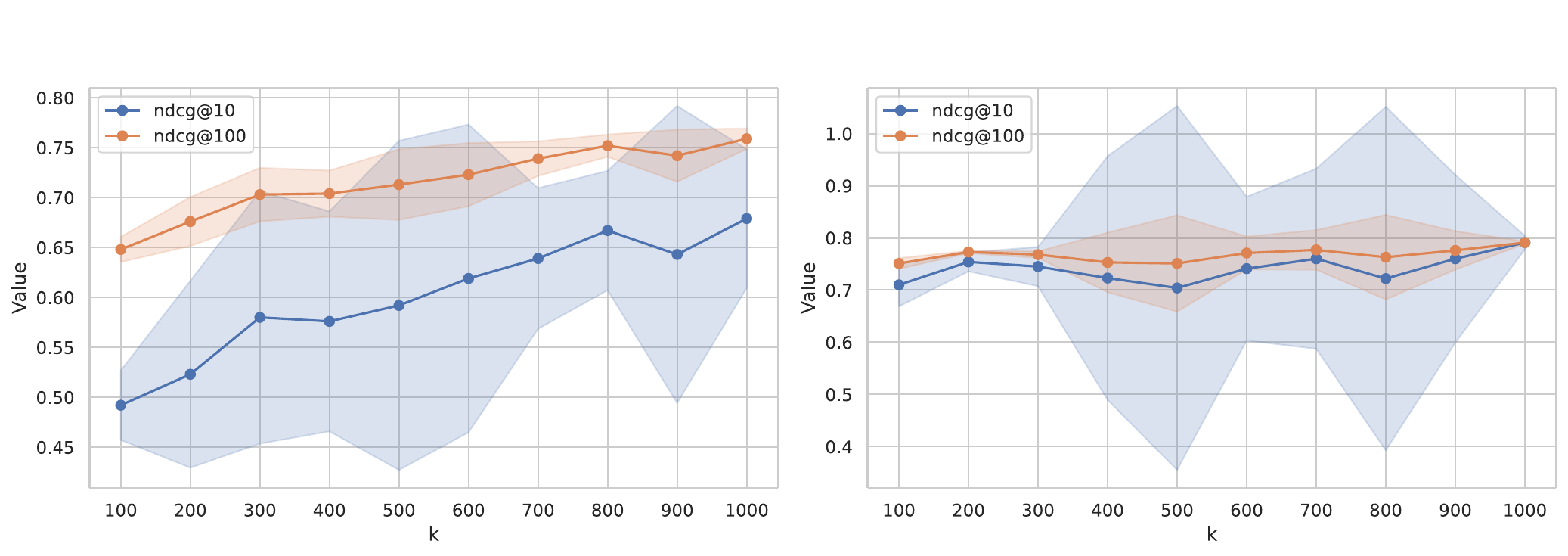}
    \caption{NDCG@10 and NDCG@100 results for the DeBERTa \textit{base} (left) and \textit{large} (right) model trained on top-k sentences.
    The shaded contour indicates the standard deviation of the scores for 5 runs.}
    \label{fig:top-k}
\end{figure*}

The results of training the DeBERTa model (\textit{base} and \textit{large} variants)
on top-$k$ sentences are given in Table \ref{tab:top-k} and Figure \ref{fig:top-k}.
For the base size, similarily to the previous experiment,  the performance of the model
increases steadily with the growing number of top sentences, until threshold of k=800 sentences per concept is reached.
Then we observe a slight drop in performance, but we also see that the variance of the results for k=900
is much higher, so the observed outcome might be just by one or few models which performed particularly bed in this setup.

The difference for NDCG@100 between k=800 and k=1000 is small (+0.7pp.).
For NDCG@10 there is +1.2 pp. difference between k=800 and k=1000. Our interpretation of these results
is that for the base model there is a tendency to
give a better score with the growing number of examples, but it seems that the plot flattens around 800 examples
per concept. It also should be noted that there is a negligible difference (+0.2 pp. for both NDCG@10 and NDCG@100) between
training on k=1000 and training on the full dataset for this setup.

\begin{table}
    \centering
    \begin{tabular}{c c c c c}
        \hline
        & \multicolumn{2}{c}{\textbf{Base}} & \multicolumn{2}{c}{\textbf{Large}} \\
        \hline
        \textbf{k} & \textbf{@10} & \textbf{@100} & \textbf{@10} & \textbf{@100} \\
        \hline
        100 & 0.492 & 0.648 & 0.710 & 0.751 \\
        200 & 0.523 & 0.676 & 0.754	& 0.773 \\
        300 & 0.580 & 0.703 & 0.745	& 0.768\\
        400 & 0.576 & 0.704 & 0.723	& 0.753\\
        500 & 0.592 & 0.713 & 0.704	& 0.751\\
        600 & 0.619 & 0.723 & 0.741	& 0.771\\
        700 & 0.639 & 0.739 & 0.760	& 0.777\\
        800 & 0.667 & 0.752 & 0.722	& 0.763\\
        900 & 0.643 & 0.742 & 0.760	& 0.776\\
        1000 & 0.679 & 0.759& \textbf{0.791} & \textbf{0.791} \\
        \hline
        full & \textbf{0.681} & \textbf{0.761}& 0.790 & 0.786 \\
    \end{tabular}
    \caption{NDCG scores @10 and @100 of training a DeBERTa (v. 3) \textit{base} and \textit{large} models on the top-$k$ sentences.}
    \label{tab:top-k}
\end{table}

The comparison between the two experiments yields the following observation.
For the base model trained without sorting we observe a huge discrepancy
between the NDCG@10 and NDCG@100 metrics, meaning that the top results
would be much worse for the base model in that setup. The gap between
these metrics is in the range 15--18 pp. gap for the randomly selected sentences,
while for the sorted sentences we observed -- 8--15 pp. gap. Moreover for the the dataset
with 1000 examples in this experiment there is a marginal (0.2 pp for both
metrics) differences compared to the full dataset. For the random version
we observe a 13 pp. difference for NDCG@10 and a 6 pp. difference for NDCG@100 between
k=1000 examples and the full dataset. The conclusion is that for the base model it is much better
to sort the sentences according to some model first, since otherwise we can
expect a huge performance reduction.

For the large model there is no such trend -- the performance fluctuates in a narrow
range for most of the settings. It grows from 100 to 200 sentences, than it falls from
200 to 500, then increases until 700 sentences, falls for 800, and grows until
1000 sentences are reached. The peak at 200  sentences is only marginally
worse than the training on the full dataset (0.7 pp. for NDCG@10 and 1.3 pp. for NDCG@100).
The results for k=1000 are better than the results of training on the full dataset
(0.1 pp. for NDCG@10 and 0.5 pp. for NDCG@100). It should be noted that the drops in
observed performance might be caused by one or several bad trainings for k=400, k=500 and k=800, since
we can observe a huge standard deviation for these values of $k$.

From the second experiment we can conclude that a base model observes steady improvements
with the growing number of examples, but it is not sensible to train the model with
the full dataset, since there is practically no difference between training with k=1000 and full
dataset. For the larger model there is no such trend and a very good performance can be
obtained with just 200 examples.

For both models we can clearly state that there is no reason to annotate
all sentences containing the legal concepts in question. Looking at the
plots it seems that 1000 sentences per concept (for the base model)
and even as few as 200 sentences per concept (for the large model) are enough
to obtain results that are only marginally worse than those obtained when annotating the
full dataset. It is thus apparent that the manual annotation effort might be
substantially reduced when only a subset of the results is annotated.

In fact if we only have a budget for 200 sentences per concept, then
there is no difference in the obtained performance between the randomly picked
sentences and the sorted ones (approx. 75\% NDCG@10 and 77\% NDCG@100).
So the setup with random sentences is the preferred one, since we don't need an existing
model to intially sort these sentences. If we want to improve
the results a bit, we can target 500 sentences per concept, when for the random version
we gain 7 pp. for NDCG@10 and 3 pp. for NDCG@100 (compared to the setup with the sorted sentences).
Using the sorted sentences with the large model only makes sense if we want to annotate up to
1000 examples for each concept, but we will only gain 1.2 pp. for NDCG@10 in that setup.

To sum up the results of this experiment, we see that for the base model
it makes much more sense to use the approach with sentence sorting, while for the
large model we will not gain much from it. To the contrary we can observe a much more
stable performance improvement, when we draw the samples randomly with the large model.
Since computing power is getting cheaper, there exist a number of efficient
training techniques and the second setup does not require a pre-existing model
for sorting the sentences, the recommended approach is to use a larger model
without sorting, allowing us to select from a range of thresholds depending
on our budget.

If we still want to train a base model, since we are concerned with the costs when
the model is deployed, we could follow the following scenario. First we should
randomly annotate 500--1000 random examples for each concept and train a large model
on that dataset. The we should use that model to sort the full set of sentences.
Then we should pick up to 1000 sentences for each concept and annotate the sentences
that were not yet annotated and we should use such a dataset to fine-tune the base model.
According to the experiments conducted so far, this should give us performance
similar to the setup when we annotated the full dataset.

\subsection{RQ3: Do we need the manual annotation at all?}

To answer the third research question we have followed the approach
presented in \cite{savelka2023can}, where the authors checked if a large language model
(GPT-4 in that case) is able to provide annotation of a good quality. The authors have found
out that the annotations provided by GPT-4 are of a medium quality -- somewhere in the middle between
the top-performing and the worst-performing human annotators. That experiment was limited in its scope
-- the authors wanted to reduce the cost of using OpenAI API, so they have only annotated automatically
256 sentences. As a result they were not able to compute the metrics used to quantify the sorting
of the sentences.

In RQ3 we have introduced the following changes in the experimental setting.
First of all, to reduce the cost of the experiment and at the same time to check if the
open source models can be a good alternative to closed models like GPT-4, we
have tested Qwen 2.5 with 72 billion parameters \cite{qwen2, qwen2.5} in the
instruct version. Since, according to our tests, the version of Qwen uploaded to HuggingFace
is invalid, i.e. it lacks definition of some tokens used for instruction fine
tuning, we have used \texttt{unsloth/Qwen2.5-72B-Instruct}. This is
an exact copy of the original model with the missing tokens included.
Secondly, since we didn't have to pay for the API we have annotated the full
test subset of the dataset (more than 11 thousand sentences).

\cite{savelka2023can} have tested two variants of the prompt used to obtain the labels.
We have followed this setup and tested exactly the same change in the prompt, which is
concerned with the definition of the certain value.

To obtain the results with optimized generation techniques (which reduce the computational
time), we have used vLLM library \cite{kwon2023efficient}. This library uses KV cache \cite{pope2022efficiently}
and prefix caching \cite{ye2024chunk} which were both turned on during the inference.
Qwen is a generative model and we have used it as such, i.e. we have not replaced the head of the model
to construct a classification network. This is not optimal, since the model predicts all values
appearing in the model's dictionary and uses autoregressive generation to provide a piece of text.
Since the only generated strings we care for are the strings \textit{no value}, \textit{potential value}, \textit{certain
value} and \textit{high value}, we have applied guided decoding \cite{beurerkellner2024guiding},
to limit the outputs of the model to include only these strings. All these techniques
contributed greatly to improving the performance of the inference and we were able
to compute the labels for the test sentences (11 thousand)
in less than 18 minutes on a node with 4 x GH200 superchips with a H100 96GB GPU.

The problem of label prediction in \cite{savelka2023can} is posed as a text classification
task. But these labels are later used to sort the sentences in order to present to the end
user those sentences that are the most valuable according to the model.
So, besides just predicting the label of the sentence, we obtained
the sentence's score by computing the probability of each valid first token associated with a
specific label and then computing the weighted sum with the labels' values mapped to numbers
(0 for no value, 1 for potential value, 2 for certain value and 3 for high value).
This allowed us to sort all sentences for a given concept according to that score and for
computing the NDCG metric for the full test set.

\begin{table}[h]
    \centering
    \begin{tabular}{c c c c c}
        \hline
       \textbf{Prompt} & \textbf{Accuracy} & \textbf{F1} & \textbf{NDCG@10} & \textbf{NDCG@100} \\
        \hline
    original  & 0.51 & 0.51 & \textbf{0.777} & \textbf{0.853} \\
    improved  & \textbf{0.54} & \textbf{0.56} & 0.766 & 0.848 \\
    \hline
    \end{tabular}
    \caption{Accuracy, (weighted) F1 and NDCG@10 and NDCG@100  scores of predicting the label by Qwen2.5-Instruct 72B on the test subset of the statutory interpretation dataset.}
    \label{tab:llm}
\end{table}

The results of this experiment are given in Table \ref{tab:llm}.
The table presents two prompts from the cited research -- the direct conversion
of the guidelines and a corrected version with an improved definition of the \textit{certain
value} label. We have used the variant without explanation and without batched
prediction. We have also employed the few-shot prompting technique by supplementing
the prompt with four examples taken randomly from the training set, one for each value of
relevance.

Regarding the obtained accuracy and F1 score -- they are pretty similar
to those obtained with GPT-4 in \cite{savelka2023can}. For the unmodified prompt it
was 0.51 and 0.53 in the
original research and we have obtained 0.51 and 0.51 (-2 pp.).
For the improved prompt it was 0.55 and 0.57 and we have obtained 0.54 (-1 pp.) and
0.56 (-1 pp.) with the Qwen model. Thus the first outcome is that currently a moderately sized (72 billion),
state-of-the-art open source model Qwen 2.5 obtains results very similar to those
of GPT-4 in the statutory interpretation task. Still we have to remember that the original research was conducted only
for a small subset of the sentences, while we have verified the results on the the full testing set.

Yet the second result is much more interesting, i.e. the NDCG scores obtained with
the prompts. The model with the original prompt achieves 0.777 NDCG@10 and 0.853 NDCG@100 while
the model with the improved prompt achieves 0.766 (-1.1 pp.) and 0.848 (-0.7 pp.) respectively.
This outcome is interesting since the accuracy and F1 scores are better for the improved prompt.

Comparing the results achievable with Qwen and manual annotation we observe that for the base model
in all setups we can achieve better results with the LLM, rather than with the manual annotation.
The best NDCG@10 score for the base model was 0.681 and it was 0.761 for NDCG@100. With the LLM
we obtain 9.6 pp. better results for NDCG@10 and 10.8 pp. better results for NDCG@100.
The best results for the  large model are 0.791 for NDCG@10 and 0.791 for NDCG@100 (for the scenario with
sorted sentences), so if we are
very much concerned with the first metric, the manual annotation of up to 1000 sentences for each
concept will give us 1.4 pp better results. For the scenario with a random sample of sentences, the difference is
negligible (0.2 pp. for NDCG@10 with 1000 sentences). Rarely such an improvement will justify the cost
of annotation. We have to also observe the fact that none of the models trained on manual annotation
achieved NDCG@100 score better than the automatic annotation with the help of Qwen.

The only concern when using the model is the computational cost of annotation. A node with 4 x GH200
chips is very expensive. Still services such as Lambda labs rent 1 GH200 for 3.32 \$/h, so the cost of annotating
even hundreds of thousands of sentences with the help of the model should be very small.

To conclude the outcome of the last experiment we state that it is sufficient for the statutory
interpretation task to use an LLM such as Qwen 2.5 with 72 billion parameters.
There is very low chance that the model trained on  the manual annotation of the dataset will yield better results
with respect to the NDCG scores, at least if we stick to fine-tuning of models with sizes and performance similar
to DeBERTa.

\section{Conclusions}

In this paper, we have attempted to optimize the laborious process of manually annotating legal texts for the extraction of statutory interpretation.
We have presented three experiments which have characterized the process along various dimensions, such as the number of annotated examples, application of a form of active learning and application of an LLM for automatic annotation. 

The results of these experiments are allowing us to provide some initial guidelines for SI extraction tasks.
First, we have seen that annotating all the sentences brings only a minor improvement over annotating a certain 
amount of them. Our experiments have shown that in general, very little gain is achieved above 1000 examples.
This highlights the potential for substantial cost reduction in annotation efforts, aligning with findings from Ein-Dor et al. \cite{ein-dor2020active} on annotation efficiency.

We have then seen that when using the base model,  active learning can help improve the results. 
On the other hand, when using a large enough model, a random choice of $k$ examples suffices. 
This result has a further interesting conclusion. One can combine a large model for training the ranking function
and then train a base model for classification of sentences based on annotation set sorted according to
the larger model. According to the second experiment presented, this should yield a much better model with
allocation of smaller resources. This supports the use of model-driven selection pipelines, as proposed in Gray et al. \cite{gray2023can} and Savelka et al. \cite{10.1145/3462757.3466149}.

Still, if we can afford to use an LLM for the annotation there will be only rare cases when we can achieve better
results with the manual annotation. In all the experiments conducted, if our primary focus is on NDCG@10 and we have the resources to annotate up to 1,000 examples for each legal concept, we can develop a model that outperforms the LLM. At the same time, we expect that the difference will be very tiny in that scenario.
So for most of the cases \textbf{we recommend doing a full automated annotation of the examples thanks to an LLM} following
the approach presented in this paper.

Does it mean that we do not have to do any manual work in order to obtain a good example of statutory interpretation?
The answer is no, because we still have to prepare the annotation guidelines (the prompt for the model) and we need 
to test the available models on some ground truth. We do not expect that this ground truth annotation should be
done automatically. But such a dataset can be rather tiny with respect to the number of examples per concept (probably
tens) in order to obtain confident performance estimates.

Nevertheless, there is still a large number of open questions that we seek to investigate in future research.
First of all we have no tried to fine-tune an LLM for the classification task. It might be the case
that such a model obtains results far better than RoBERTa and for this scenario we still need 
manually annotated training examples. It is an open question how many such examples are necessary
to beat a model based only on the prompts.

The second open question concerns the optimization of the prompt that is used by the LLM.
Thanks to tools such as DSPy \cite{khattab2023dspy} this process can be automated and since the cost of using
open source LLMs is small, we could follow a number of experiments to find out what results are
achievable with the help of that technique. 

We have also limited ourselves to test just one open source LLM -- namely Qwen, while there exist 
a large number of such models. It would be very instructive to compere these models on the
statutory interpretation task, with careful considerations balancing the performance
of the models and the cost of inference. 

Another question is with regards to the distribution of the classes. In this paper, we have assumed that 
by taking first more relevant examples, we are producing a more accurate model. This assumption should be verified via 
further experiments.

This research is a part of a wider one to make the law more accessible to lay people. As such, statutory interpretation 
extraction plays a crucial role and we are planning to answer the above, and further, questions in the future research.

\section{Acknowledgements}

Supported by the Polish National Centre for Research and Development – Pollux Program under Grant WM/POLLUX11/5/2023 titled ,,Examples based AI Legal Guidance''.

We gratefully acknowledge Polish high-performance computing infrastructure PLGrid (HPC Center: ACK Cyfronet AGH) for providing computer facilities and support within computational grant no. PLG/2024/017168.

\bibliographystyle{acm}
\bibliography{0-main}

\begin{thebibliography}{10}

\bibitem{ashley2013information}
{\sc Ashley, K., and Walker, V.}
\newblock From information retrieval (ir) to argument retrieval (ar) for legal
  cases: Report on a baseline study.
\newblock {\em Frontiers in Artificial Intelligence and Applications 259\/} (01
  2013), 29--38.

\bibitem{beurerkellner2024guiding}
{\sc Beurer-Kellner, L., Fischer, M., and Vechev, M.}
\newblock Guiding llms the right way: Fast, non-invasive constrained
  generation, 2024.

\bibitem{bommarito2023gpt}
{\sc Bommarito, M.~J., and Katz, D.}
\newblock 2022.
\newblock {\em arXiv Preprint\/}.

\bibitem{chalkidis2020legalbert}
{\sc Chalkidis, I., Fergadiotis, M., Malakasiotis, P., Aletras, N., and
  Androutsopoulos, I.}
\newblock {LEGAL}-{BERT}: The muppets straight out of law school.
\newblock In {\em Findings of the Association for Computational Linguistics:
  EMNLP 2020\/} (Online, Nov. 2020), T.~Cohn, Y.~He, and Y.~Liu, Eds.,
  Association for Computational Linguistics, pp.~2898--2904.

\bibitem{bge2024chen}
{\sc Chen, J., Xiao, S., Zhang, P., Luo, K., Lian, D., and Liu, Z.}
\newblock Bge m3-embedding: Multi-lingual, multi-functionality,
  multi-granularity text embeddings through self-knowledge distillation, 2024.

\bibitem{DBLP:journals/corr/abs-2411-07739}
{\sc de~Oliveira~Lima, J.~A.}
\newblock Unlocking legal knowledge with multi-layered embedding-based
  retrieval.
\newblock {\em CoRR abs/2411.07739\/} (2024).

\bibitem{ein-dor2020active}
{\sc Ein-Dor, L., Halfon, A., Gera, A., Shnarch, E., Dankin, L., Choshen, L.,
  Danilevsky, M., Aharonov, R., Katz, Y., and Slonim, N.}
\newblock {A}ctive {L}earning for {BERT}: {A}n {E}mpirical {S}tudy.
\newblock In {\em Proceedings of the 2020 Conference on Empirical Methods in
  Natural Language Processing (EMNLP)\/} (Online, Nov. 2020), B.~Webber,
  T.~Cohn, Y.~He, and Y.~Liu, Eds., Association for Computational Linguistics,
  pp.~7949--7962.

\bibitem{gray2023can}
{\sc Gray, M., Savelka, J., Oliver, W., and Ashley, K.}
\newblock Can gpt alleviate the burden of annotation?
\newblock In {\em Legal Knowledge and Information Systems}. IOS Press, 2023,
  pp.~157--166.

\bibitem{he2021debertav3}
{\sc He, P., Gao, J., and Chen, W.}
\newblock Debertav3: Improving deberta using electra-style pre-training with
  gradient-disentangled embedding sharing, 2021.

\bibitem{jarvelin2002cumulated}
{\sc J{\"a}rvelin, K., and Kek{\"a}l{\"a}inen, J.}
\newblock Cumulated gain-based evaluation of ir techniques.
\newblock {\em ACM Transactions on Information Systems (TOIS) 20}, 4 (2002),
  422--446.

\bibitem{Katz2024gpt4}
{\sc Katz, D., Bommarito, M., Gao, S., and Arredondo, P.}
\newblock Gpt-4 passes the bar exam.
\newblock {\em Philosophical Transactions of the Royal Society A 382\/} (02
  2024).

\bibitem{khattab2023dspy}
{\sc Khattab, O., Singhvi, A., Maheshwari, P., Zhang, Z., Santhanam, K.,
  Vardhamanan, S., Haq, S., Sharma, A., Joshi, T.~T., Moazam, H., Miller, H.,
  Zaharia, M., and Potts, C.}
\newblock Dspy: Compiling declarative language model calls into self-improving
  pipelines, 2023.

\bibitem{kwon2023efficient}
{\sc Kwon, W., Li, Z., Zhuang, S., Sheng, Y., Zheng, L., Yu, C.~H., Gonzalez,
  J.~E., Zhang, H., and Stoica, I.}
\newblock Efficient memory management for large language model serving with
  pagedattention.
\newblock In {\em Proceedings of the ACM SIGOPS 29th Symposium on Operating
  Systems Principles\/} (2023).

\bibitem{unknown}
{\sc Luo, K., Huang, Q., Jiang, C., and Feng, Y.}
\newblock Automating legal concept interpretation with llms: Retrieval,
  generation, and evaluation, 01 2025.

\bibitem{niklaus2022empirical}
{\sc Niklaus, J., St{\"u}rmer, M., and Chalkidis, I.}
\newblock An empirical study on cross-x transfer for legal judgment prediction.
\newblock {\em arXiv preprint arXiv:2209.12325\/} (2022).

\bibitem{pope2022efficiently}
{\sc Pope, R., Douglas, S., Chowdhery, A., Devlin, J., Bradbury, J., Levskaya,
  A., Heek, J., Xiao, K., Agrawal, S., and Dean, J.}
\newblock Efficiently scaling transformer inference, 2022.

\bibitem{savelka2020discovering}
{\sc Savelka, J.}
\newblock {\em Discovering sentences for argumentation about the meaning of
  statutory terms}.
\newblock PhD thesis, University of Pittsburgh, 2020.

\bibitem{vsavelka2015transfer}
{\sc {\v{S}}avelka, J., and Ashley, K.~D.}
\newblock Transfer of predictive models for classification of statutory texts
  in multi-jurisdictional settings.
\newblock In {\em Proceedings of the 15th International Conference on
  Artificial Intelligence and Law\/} (2015), pp.~216--220.

\bibitem{savelka2021discovering}
{\sc Savelka, J., and Ashley, K.~D.}
\newblock Discovering explanatory sentences in legal case decisions using
  pre-trained language models.
\newblock {\em In: Findings of the Association for Computational Linguistics:
  EMNLP 2021. pp. 4273–4283\/} (2021).

\bibitem{vsavelka2022legal}
{\sc {\v{S}}avelka, J., and Ashley, K.~D.}
\newblock Legal information retrieval for understanding statutory terms.
\newblock {\em Artificial Intelligence and Law\/} (2022), 1--45.

\bibitem{savelka2023can}
{\sc Savelka, J., Ashley, K.~D., Gray, M.~A., Westermann, H., and Xu, H.}
\newblock Can gpt-4 support analysis of textual data in tasks requiring highly
  specialized domain expertise?
\newblock {\em arXiv preprint arXiv:2306.13906\/} (2023).

\bibitem{savelka2021cross}
{\sc Savelka, J., Westermann, H., and Benyekhlef, K.}
\newblock Cross-domain generalization and knowledge transfer in transformers
  trained on legal data.
\newblock {\em arXiv preprint arXiv:2112.07870\/} (2021).

\bibitem{10.1145/3462757.3466149}
{\sc Savelka, J., Westermann, H., Benyekhlef, K., Alexander, C.~S., Grant,
  J.~C., Amariles, D.~R., Hamdani, R.~E., Mee\`{u}s, S., Troussel, A.,
  Araszkiewicz, M., Ashley, K.~D., Ashley, A., Branting, K., Falduti, M.,
  Grabmair, M., Hara\v{s}ta, J., Novotn\'{a}, T., Tippett, E., and Johnson, S.}
\newblock Lex rosetta: transfer of predictive models across languages,
  jurisdictions, and legal domains.
\newblock In {\em Proceedings of the Eighteenth International Conference on
  Artificial Intelligence and Law\/} (New York, NY, USA, 2021), ICAIL '21,
  Association for Computing Machinery, p.~129–138.

\bibitem{smywinski2024enhancing}
{\sc Smywi{\'n}ski-Pohl, A., and Libal, T.}
\newblock Enhancing legal argument retrieval with optimized language model
  techniques.
\newblock In {\em JSAI International Symposium on Artificial Intelligence\/}
  (2024), Springer, pp.~93--108.

\bibitem{qwen2.5}
{\sc Team, Q.}
\newblock Qwen2.5: A party of foundation models, September 2024.

\bibitem{tyss2024beyond}
{\sc Tyss, S., Venkatkrishna, V., Ghosh, S., and Grabmair, M.}
\newblock Beyond borders: Investigating cross-jurisdiction transfer in legal
  case summarization.
\newblock In {\em Proceedings of the 2024 Conference of the North American
  Chapter of the Association for Computational Linguistics: Human Language
  Technologies (Volume 1: Long Papers)\/} (2024), pp.~4136--4150.

\bibitem{qwen2}
{\sc Yang, A., Yang, B., Hui, B., Zheng, B., Yu, B., Zhou, C., Li, C., Li, C.,
  Liu, D., Huang, F., Dong, G., Wei, H., Lin, H., Tang, J., Wang, J., Yang, J.,
  Tu, J., Zhang, J., Ma, J., Xu, J., Zhou, J., Bai, J., He, J., Lin, J., Dang,
  K., Lu, K., Chen, K., Yang, K., Li, M., Xue, M., Ni, N., Zhang, P., Wang, P.,
  Peng, R., Men, R., Gao, R., Lin, R., Wang, S., Bai, S., Tan, S., Zhu, T., Li,
  T., Liu, T., Ge, W., Deng, X., Zhou, X., Ren, X., Zhang, X., Wei, X., Ren,
  X., Fan, Y., Yao, Y., Zhang, Y., Wan, Y., Chu, Y., Liu, Y., Cui, Z., Zhang,
  Z., and Fan, Z.}
\newblock Qwen2 technical report.
\newblock {\em arXiv preprint arXiv:2407.10671\/} (2024).

\bibitem{ye2024chunk}
{\sc Ye, L., Tao, Z., Huang, Y., and Li, Y.}
\newblock Chunkattention: Efficient self-attention with prefix-aware kv cache
  and two-phase partition, 2024.

\bibitem{zheng2021pretraining}
{\sc Zheng, L., Guha, N., Anderson, B., Henderson, P., and Ho, D.}
\newblock When does pretraining help?: Assessing self-supervised learning for
  law and the casehold dataset of 53,000+ legal holdings.
\newblock In {\em Proceedings of the 18th International Conference on
  Artificial Intelligence and Law, ICAIL 2021\/} (June 2021), Proceedings of
  the 18th International Conference on Artificial Intelligence and Law, ICAIL
  2021, Association for Computing Machinery, Inc, pp.~159--168.
\newblock Publisher Copyright: {\textcopyright} 2021 Owner/Author.; 18th
  International Conference on Artificial Intelligence and Law, ICAIL 2021 ;
  Conference date: 21-06-2021 Through 25-06-2021.

\end{thebibliography}

\end{document}